\documentclass[11pt]{airesearch}
\usepackage[T1]{fontenc}
\usepackage{XCharter}

\usepackage{algorithm}
\usepackage{algpseudocode}  

\ifdefined\final
  \usepackage[disable]{todonotes}
\else
  \usepackage[textsize=tiny]{todonotes}
\fi

\usepackage{tikz}
\usetikzlibrary{positioning, fit, calc, shapes.geometric, arrows.meta, shadows, shadows.blur, backgrounds}

\definecolor{appleBlue}{RGB}{0, 122, 255}
\definecolor{appleLightBlue}{RGB}{235, 243, 255}
\definecolor{appleGreen}{RGB}{52, 199, 89}
\definecolor{appleLightGreen}{RGB}{238, 249, 239}
\definecolor{appleOrange}{RGB}{255, 149, 0}
\definecolor{appleLightOrange}{RGB}{255, 247, 235}
\definecolor{appleGray}{RGB}{142, 142, 147}
\definecolor{appleLightGray}{RGB}{242, 242, 247}
\definecolor{appleDark}{RGB}{28, 28, 30}

\title{\huge \bfseries Deep Delta Learning}
\author{
    \textbf{Yifan Zhang}$^{1}$~~~~\textbf{Yifeng Liu}$^{2}$~~~~\textbf{Mengdi Wang}$^{1}$~~~~\textbf{Quanquan Gu}$^2$\\[1.5mm]
    $^1$Princeton University~~~~$^2$University of California, Los Angeles\\[0.5mm]
    \texttt{yifzhang@princeton.edu}
}
\date{}

\begin{document}
\maketitle


\begin{abstract}
We introduce \textbf{Deep Delta Learning (DDL)}, which replaces the additive residual update of the Transformer with the delta rule applied across depth. A standard sublayer adds its output to the residual stream regardless of what the stream already holds. In DDL, each attention or MLP sublayer selects a direction, reads the state along it, and writes a gated rank-1 correction proportional to the error between a learned target and that readout. A single gate interpolates between skipping the sublayer and overwriting the selected component, and every orthogonal direction stays on the identity path. The same rule extends to a residual stream with several value channels per feature, which enlarges the persistent state of the model while attention and MLP widths stay fixed. In decoder-only language models at \mbox{GPT-2} small and medium scales, each pretrained on 49.15B FineWeb-Edu tokens, every DDL variant reaches lower validation loss and higher mean one-shot accuracy on eight downstream tasks than the additive baseline, with fewer than 0.3\% additional parameters. 
\end{abstract}

\projectpage{https://github.com/yifanzhang-pro/deep-delta-learning}

\section{Introduction}

\begingroup
\definecolor{ddlInk}{HTML}{20344A}
\definecolor{ddlLine}{HTML}{607A93}
\definecolor{ddlBorder}{HTML}{AABBCB}
\definecolor{ddlSurface}{HTML}{EFF4F9}
\definecolor{ddlAccent}{HTML}{275D8C}
\definecolor{ddlAccentLine}{HTML}{315F89}
\definecolor{ddlMuted}{HTML}{4C6882}
\tikzset{
    ddl base/.style={font=\sffamily\fontsize{9}{11}\selectfont,
        text=ddlInk, line width=0.65pt, inner sep=4pt, align=center},
    ddl box/.style={ddl base, rounded corners=2pt, draw=ddlBorder,
        fill=ddlSurface!45, minimum height=0.68cm},
    ddl stage/.style={ddl box, minimum width=3.2cm, minimum height=1.0cm},
    ddl arrow/.style={draw=ddlLine, line width=0.8pt,
        -{Stealth[length=1.8mm,width=1.25mm]}, rounded corners=2pt},
    ddl skip/.style={ddl arrow, draw=ddlInk, line width=1.0pt},
    ddl note/.style={ddl base, font=\sffamily\fontsize{8}{10}\selectfont},
}
\begin{figure}[t]
    \centering
    \begin{subfigure}[t]{0.485\textwidth}
        \centering
        \begin{tikzpicture}[ddl base]
            \path[use as bounding box] (-0.35,0.55) rectangle (6.25,-9.05);
            \node[ddl box, minimum width=2.3cm] (state) at (2.65,0)
                {$\Xb_l\in\RR^{d\times d_v}$};
            \node[ddl stage, draw=ddlBorder, fill=ddlSurface] (read) at (2.65,-1.65)
                {\textbf{Read}\\$\mathbf{r}_l=\kb_l^\top\Xb_l$};
            \node[ddl stage, draw=ddlAccentLine, fill=ddlAccent!14!ddlSurface] (compare) at (2.65,-3.2)
                {\textbf{Compare}\\$\Delta_l=\vb_l^\top-\mathbf{r}_l$};
            \node[ddl stage, draw=ddlAccentLine, fill=ddlAccent, text=white] (write) at (2.65,-4.75)
                {\textbf{Write}\\$\mathbf{U}_l=\beta_l\kb_l\Delta_l$};
            \node[ddl box, circle, fill=white, draw=ddlInk,
                minimum size=0.55cm, inner sep=0pt] (add) at (2.65,-6.35) {$+$};
            \node[ddl box, minimum width=3.2cm] (out) at (2.65,-7.5)
                {$\Xb_{l+1}=\Xb_l+\mathbf{U}_l$};

            \coordinate (fork) at (2.65,-0.75);
            \draw[ddl arrow] (state) -- (read);
            \draw[ddl arrow] (read) -- (compare);
            \draw[ddl arrow] (compare) -- (write);
            \draw[ddl arrow] (write) -- (add);
            \draw[ddl arrow] (add) -- (out);
            \draw[ddl skip] (fork) -- (0.15,-0.75) |- (add.west);
            \fill[ddlInk] (fork) circle (1.2pt);
            \node[ddl note, rotate=90, fill=white, inner sep=2pt]
                at (0.15,-3.45) {Identity path};

            \node[ddl note, text=ddlMuted] (key) at (5.35,-1.65)
                {Direction\\$\kb_l$};
            \node[ddl note, text=ddlMuted] (value) at (5.35,-3.2)
                {Target\\$\vb_l^\top$};
            \node[ddl note, text=ddlInk] (gate) at (5.35,-4.75)
                {Gate $\beta_l$\\Direction $\kb_l$};
            \draw[ddl arrow, draw=ddlLine] (key.west) -- (read.east);
            \draw[ddl arrow, draw=ddlLine] (value.west) -- (compare.east);
            \draw[ddl arrow, draw=ddlAccentLine] (gate.west) -- (write.east);

            \node[ddl note, text width=5.9cm] at (2.65,-8.6)
                {$\|\kb_l\|_2=1$\quad\textbullet\quad shared erase/write gate\\
                 $\beta_l=0$: identity\qquad $\beta_l=1$: target match};
        \end{tikzpicture}
        \caption{The residual rewrite.}
        \label{fig:deep_delta_block}
    \end{subfigure}
    \hfill
    \begin{subfigure}[t]{0.485\textwidth}
        \centering
        \begin{tikzpicture}[ddl base]
            \path[use as bounding box] (-0.35,0.55) rectangle (6.25,-9.05);
            \node[ddl box, minimum width=2.3cm] (state) at (2.65,0)
                {$\Xb_l\in\RR^{d\times d_v}$};
            \node[ddl stage, minimum height=0.85cm] (compress) at (2.65,-1.45)
                {\textbf{Compress}\\$\xb_l^{\mathrm{in}}=\mathcal{C}_l(\Xb_l)$};
            \node[ddl stage, minimum height=0.85cm] (norm) at (2.65,-2.7)
                {\textbf{Normalize}\\$\mathbf{c}_l=\operatorname{RMSNorm}(\xb_l^{\mathrm{in}})$};
            \node[ddl stage, draw=ddlInk, fill=ddlInk, text=white] (process) at (2.65,-4.2)
                {\textbf{Attention / MLP}\\$\mathbf{h}_l=\Fb_l(\mathbf{c}_l)$};
            \node[ddl stage, minimum height=0.75cm,
                draw=ddlBorder, fill=ddlSurface] (key) at (2.65,-5.55)
                {$\kb_l=\mathbf{h}_l/\|\mathbf{h}_l\|_2$};
            \node[ddl box, minimum width=1.25cm, minimum height=1.2cm,
                draw=ddlAccentLine, fill=ddlAccent!14!ddlSurface, font=\sffamily\fontsize{8}{10}\selectfont]
                (heads) at (5.45,-4.2) {\textbf{Value / gate}\\$\vb_l,\beta_l$};
            \node[ddl stage, minimum width=4.6cm,
                draw=ddlAccentLine, fill=ddlAccent, text=white] (rewrite) at (2.65,-6.85)
                {\textbf{Rewrite}\; (a)\\$\Xb_{l+1}=\Xb_l+\beta_l\kb_l\Delta_l$};
            \node[ddl box, minimum width=2.3cm] (out) at (2.65,-8.0)
                {$\Xb_{l+1}\in\RR^{d\times d_v}$};

            \coordinate (fork) at (2.65,-0.65);
            \coordinate (input) at (2.65,-2.075);
            \coordinate (context) at (2.65,-3.4);
            \draw[ddl arrow] (state) -- (compress);
            \draw[ddl arrow] (compress) -- (norm);
            \draw[ddl arrow] (norm) -- (process);
            \draw[ddl arrow, draw=ddlLine] (process) -- (key);
            \draw[ddl arrow, draw=ddlAccentLine] (key) -- (rewrite);
            \draw[ddl arrow] (rewrite) -- (out);
            \draw[ddl skip] (fork) -- (0.0,-0.65) |- (rewrite.west);
            \fill[ddlInk] (fork) circle (1.2pt);
            \node[ddl note, rotate=90, fill=white, inner sep=2pt]
                at (0.0,-3.45) {Persistent residual state};
            \draw[ddl arrow, draw=ddlAccentLine] (input) -| ([xshift=3mm]heads.north);
            \fill[ddlAccentLine] (input) circle (1.2pt);
            \node[ddl note, text=ddlMuted, anchor=south, inner sep=1pt] at (4.85,-2.075)
                {$\xb_l^{\mathrm{in}}$};
            \draw[ddl arrow, draw=ddlAccentLine] (context) -| ([xshift=-3mm]heads.north);
            \fill[ddlAccentLine] (context) circle (1.2pt);
            \node[ddl note, text=ddlMuted, anchor=south, inner sep=1pt] at (4.6,-3.4)
                {$\mathbf{c}_l$};
            \draw[ddl arrow, draw=ddlAccentLine] (heads.south)
                -- (5.45,-6.0) -- (4.15,-6.0) -- (4.15,-6.35);

            \node[ddl note, text width=6.1cm] at (2.65,-8.75)
                {Repeat across depth\quad\textbullet\quad sublayer width stays $d$};
        \end{tikzpicture}
        \caption{A DDL Transformer sublayer.}
        \label{fig:ddl_transformer_structure}
    \end{subfigure}
    \caption{\textbf{Deep Delta Learning overview.} (a) Read the selected residual content, compare it with a target, and add a gated rank-1 correction to the identity path. (b) A width-$d$ sublayer produces the direction; lightweight branches produce the target from the sublayer input $\xb_l^{\mathrm{in}}$ and the gate from the normalized context $\mathbf{c}_l$. The expanded residual persists across sublayers. Batch and sequence axes are omitted.}
    \label{fig:ddl_overview}
\end{figure}
\endgroup

Residual connections give every layer an identity path, which keeps deep networks trainable~\citep{he2016deep}. In a Transformer, the residual stream $\xb_l\in\RR^{B\times T\times d}$ (batch size $B$, sequence length $T$, width $d$) also serves as the working memory of the model: each attention or MLP sublayer reads it and adds its output back, $\xb_{l+1}=\xb_l+\Fb_l(\xb_l)$. Addition can only accumulate: to revise a feature written by an earlier layer, a sublayer must infer the current value and emit the difference on its own, because the update never consults what the stream holds along the direction being written.

Deep Delta Learning (\textbf{DDL}) builds this comparison into the residual update. The delta rule~\citep{widrow1988adaptive} corrects a stored value in proportion to its error, and sequence-memory models use it to update a memory over time~\citep{schlag2021linear,yang2024parallelizing}. We apply it to the residual state $\Xb_l$ over depth,
\begin{align*}
\Xb_{l+1}=\Xb_l+\beta_l\kb_l\bigl(\vb_l^\top-\kb_l^\top\Xb_l\bigr).
\end{align*}
The unit vector $\kb_l$ points along the sublayer output, so $\kb_l^\top\Xb_l$ reads the content that the sublayer is about to revise. Two lightweight branches predict the target $\vb_l$ and the gate $\beta_l\in(0,2)$. The update moves the readout a fraction $\beta_l$ of the way to the target and leaves every direction orthogonal to $\kb_l$ unchanged: $\beta_l\to0$ skips the sublayer, $\beta_l=1$ overwrites the readout with the target, and $\beta_l>1$ overshoots it. Each sublayer therefore takes one gradient step on a squared readout error, with the gate as its step size (Figure~\ref{fig:ddl_overview}).

Because the update acts along the feature axis, it carries over to a wider state without modification. We let the residual stream hold $d_v$ value channels per feature, compress it to a width-$d$ vector at the input of each sublayer, and revise all channels with one rank-1 update. The persistent state grows by a factor of $d_v$ while attention and MLP sublayers keep width $d$.

\paragraph{Contributions.}
\begin{enumerate}[leftmargin=*, itemsep=0pt, topsep=0pt]
\item We formulate the residual update as a delta rule over depth and analyze it for a fixed direction and gate. The shortcut $\Ib-\beta_l\kb_l\kb_l^\top$ has eigenvalue $1-\beta_l$ along $\kb_l$ and $1$ on its orthogonal complement, and each sublayer contracts the readout error by $|1-\beta_l|$.
\item We extend DDL from a vector state ($d_v=1$) to a residual state $\Xb_l\in\RR^{B\times T\times d\times d_v}$ through a Compress--Process--Rewrite layer, with two learned compressors and an embedding convolution that initializes the value channels.
\item At GPT-2 small and medium scales~\citep{radford2019language}, with 49.15B FineWeb-Edu tokens per model, all five DDL variants improve on the additive baseline in validation loss and in mean one-shot accuracy over eight tasks. Expanded-state DDL lowers validation loss by up to $0.0244$ and $0.0295$ and raises one-shot accuracy by up to $0.91$ and $1.18$ points at the two scales. At the small scale it also reaches lower validation loss than Hyper-Connections~\citep{zhuhyper2025} at the same expansion rate.
\end{enumerate}

\section{Deep Delta Learning}
\label{sec:deep_delta_learning}

We derive the update for one token and layer, suppressing batch and sequence axes. The residual state is $\Xb_l\in\RR^{d\times d_v}$, where $d$ is the Transformer width and $d_v$ the number of value channels; $d_v=1$ recovers a vector state.

\subsection{Delta Residual Rewrite}

Given $\Xb_l$, three generator functions produce an unnormalized direction $\tilde{\kb}_l=\mathcal{K}_l(\Xb_l)\in\RR^d$, a target value $\vb_l=\mathcal{V}_l(\Xb_l)\in\RR^{d_v}$, and a gate $\beta_l=\mathcal{B}_l(\Xb_l)\in(0,2)$. These functions are instantiated in Section~\ref{sec:ddl_transformer}: $\mathcal{K}_l$ contains the standard sublayer, while $\mathcal{V}_l$ and $\mathcal{B}_l$ are lightweight branches on the sublayer input.

\paragraph{Read/write direction.}
We normalize the proposed direction to $\kb_l=\tilde{\kb}_l/\|\tilde{\kb}_l\|_2$, so $\|\kb_l\|_2=1$, with a small-norm guard in implementation. The row vector $\kb_l^\top\Xb_l\in\RR^{1\times d_v}$ is the current readout along the selected feature direction, and $\vb_l^\top$ is its target value.

\paragraph{Erase and write as one affine update.}
Conditioned on the generated quantities, define the direct shortcut operator $\Ab_l=\Ib-\beta_l\kb_l\kb_l^\top$. DDL applies this shortcut and writes a rank-1 target along the same direction:
\begin{equation}
\label{eq:gated_hres_out}
\Xb_{l+1}=\Ab_l\Xb_l+\beta_l\kb_l\vb_l^\top.
\end{equation}
Equivalently,
\begin{equation}
\label{eq:ddl_additive}
\Xb_{l+1}
=
\Xb_l+\beta_l\kb_l\Bigl(\vb_l^\top-\kb_l^\top\Xb_l\Bigr).
\end{equation}
The discrepancy $\vb_l^\top-\kb_l^\top\Xb_l\in\RR^{1\times d_v}$ vanishes when the selected readout already matches the target. Multiplication by $\kb_l$ confines the direct correction to $\mathrm{span}\{\kb_l\}$; directions orthogonal to $\kb_l$ remain on the identity shortcut after conditioning on the generated quantities.

\paragraph{Shared gate and exact overwrite.}
The same $\beta_l$ controls both erasure and writing. At $\beta_l=0$, Eq.~\eqref{eq:ddl_additive} gives $\Xb_{l+1}=\Xb_l$. At $\beta_l=1$, $\kb_l^\top\Xb_{l+1}=\vb_l^\top$, so the one-dimensional readout along $\kb_l$ is exactly replaced by the target. Tying the two makes $\beta_l$ an interpolation weight between the old readout and the target.

\paragraph{Target-seeking error correction.}
Eq.~\eqref{eq:ddl_additive} is one gradient step on a local regression loss over the state. With $\kb_l$ and $\vb_l$ held at their generated values, define for a candidate state $\Zb\in\RR^{d\times d_v}$
\begin{equation}
\label{eq:local_loss}
\ell_l(\Zb)=\tfrac{1}{2}\bigl\|\kb_l^\top\Zb-\vb_l^\top\bigr\|_F^2,
\qquad
\nabla_{\Zb}\ell_l(\Zb)=\kb_l\bigl(\kb_l^\top\Zb-\vb_l^\top\bigr).
\end{equation}
Then $\Xb_{l+1}=\Xb_l-\beta_l\nabla_{\Zb}\ell_l(\Xb_l)$, so the gate $\beta_l$ is the step size. The step is taken in closed form in the forward pass, and all parameters are trained on the language-model objective.

Let $\mathbf{e}_l^{\mathrm{pre}}=\kb_l^\top\Xb_l-\vb_l^\top$ and $\mathbf{e}_l^{\mathrm{post}}=\kb_l^\top\Xb_{l+1}-\vb_l^\top$. Using $\|\kb_l\|_2=1$ in Eq.~\eqref{eq:ddl_additive} gives $\mathbf{e}_l^{\mathrm{post}}=(1-\beta_l)\mathbf{e}_l^{\mathrm{pre}}$, and hence $\ell_l(\Xb_{l+1})=(1-\beta_l)^2\,\ell_l(\Xb_l)$. Therefore $0<\beta_l<2$ contracts the discrepancy in magnitude by $|1-\beta_l|$, $\beta_l=1$ removes it exactly, and $1<\beta_l<2$ changes its sign while reducing its magnitude. The contraction holds at every sublayer for the generated $\kb_l$, $\vb_l$, and $\beta_l$.

The sigmoid gate used below yields $\beta_l\in(0,2)$, the step sizes that decrease a nonzero $\ell_l$; endpoints are saturated-logit limits.

\subsection{Expanded Residual State}

For $d_v>1$, DDL separates persistent storage from sublayer compute through a \emph{Compress--Process--Rewrite} interface: compress $\Xb_l$ to a width-$d$ vector, process it with attention or an MLP, and rewrite the expanded state via Eq.~\eqref{eq:ddl_additive}. Attention keys, queries, values, and MLP hidden activations are not widened to $dd_v$.

\subsection{Spectral Analysis}

We give the spectrum of the direct shortcut for a fixed token and state, conditioned on the generated direction and gate.

\begin{proposition}[Frozen shortcut spectrum]
\label{prop:spectrum}
Let $\Ab=\Ib-\beta\kb\kb^\top$, where $\kb\in\RR^d$ is a unit vector and $\beta\in\RR$ is fixed. If $\beta\neq0$, the eigenvalues of $\Ab$ are $1$ with multiplicity $d-1$ and $1-\beta$ with multiplicity $1$. The eigenvector for $1-\beta$ is $\kb$, and the eigenspace for eigenvalue $1$ is $\kb^\perp=\{\ub\in\RR^d:\kb^\top\ub=0\}$. If $\beta=0$, then $\Ab=\Ib$ and the eigenspace for eigenvalue $1$ is all of $\RR^d$.
\end{proposition}
\begin{proof}
For any $\ub\in\kb^\perp$, $\Ab\ub=\ub-\beta\kb(\kb^\top\ub)=\ub$. Also, $\Ab\kb=\kb-\beta\kb(\kb^\top\kb)=(1-\beta)\kb$. When $\beta\neq0$, these $d$ independent eigen-directions span $\RR^d$; when $\beta=0$, the claim reduces to $\Ab=\Ib$.
\end{proof}

For any $\ub=\ub_\perp+(\kb^\top\ub)\kb$ with $\ub_\perp\in\kb^\perp$, $\Ab\ub=\ub_\perp+(1-\beta)(\kb^\top\ub)\kb$. Thus the frozen shortcut leaves $\kb^\perp$ unchanged and scales only the selected direction. For a matrix-valued residual state, the same left-multiplying operator acts on each of the $d_v$ value columns; under vectorization, the shortcut is $\Ib_{d_v}\otimes\Ab$. Since $\det\Ab=1-\beta$, the conditioned update $\Xb\mapsto\Ab\Xb+\beta\kb\vb^\top$ is an invertible affine map for every $\beta\neq1$; at $\beta=1$, $\Ab$ is the orthogonal projector onto $\kb^\perp$, and the old readout along $\kb$ is discarded.

Together with the synchronized write, the selected readout evolves as $\kb^\top\Xb_{l+1}=\vb^\top+(1-\beta)(\kb^\top\Xb_l-\vb^\top)$, the error contraction above, which yields three local regimes:
\begin{itemize}[leftmargin=*, topsep=1pt, itemsep=1pt]
\item \textbf{$\beta\approx0$: skip.} The shortcut approaches $\Ib$, the write term vanishes, and the complete update approaches the identity.
\item \textbf{$\beta\approx1$: target match.} The old readout along $\kb$ is removed and replaced by $\vb^\top$, exactly so at $\beta=1$.
\item \textbf{$\beta>1$: over-relaxed correction.} The readout crosses the target because $1-\beta<0$. At the saturated endpoint $\beta\to2$, the direct shortcut approaches the Householder reflector $\Ib-2\kb\kb^\top$; the complete affine update does not.
\end{itemize}

The analysis gives each sublayer update an explicit reading: which one-dimensional subspace is edited, what target is requested, and how strongly the discrepancy is corrected.

\subsection{Relation to DeltaNet}
\label{sec:relation_delta_nets}

DDL takes its update from the delta rule of DeltaNet~\citep{schlag2021linear}, which replaces additive accumulation in linear Transformers with a delta-rule memory update. Written with left multiplication, DeltaNet updates a memory $\Sbb_t\in\RR^{d_k\times d_v}$ over time $t$ as
\begin{equation}
\label{eq:deltanet_eq}
\Sbb_t=(\Ib-\beta_t\kb_t\kb_t^\top)\Sbb_{t-1}+\beta_t\kb_t\vb_t^\top,
\end{equation}
which is the more common update $\Mb_t=\Mb_{t-1}+\beta_t(\vb_t-\Mb_{t-1}\kb_t)\kb_t^\top$ with $\Sbb_t=\Mb_t^\top$. Eq.~\eqref{eq:deltanet_eq} and the DDL update Eq.~\eqref{eq:gated_hres_out} correspond term by term. The memory $\Sbb_t$ corresponds to the residual state $\Xb_l$, and the key dimension $d_k$ to the feature dimension $d$. Both apply the shortcut $\Ib-\beta\kb\kb^\top$, which for $\|\kb\|_2=1$ is an orthogonal projector at $\beta=1$ and a Householder reflection at $\beta=2$; DeltaNet applies it over time steps, DDL over network depth. Both write $\beta\kb\vb^\top$, so the same $\beta$ scales erasure and writing and acts as a step size, which gives the complete DDL update its identity limit at $\beta_l=0$. Reading the delta rule as one gradient step on a squared readout error is established for sequence memories~\citep{yang2024parallelizing}; Eq.~\eqref{eq:local_loss} is the same loss with the residual state in place of the memory.

\section{DDL Transformer}
\label{sec:ddl_transformer}

We evaluate DDL in decoder-only Transformer language models with pre-norm RMSNorm, RoPE multi-head attention, and SwiGLU MLPs. DDL changes the residual interface while preserving the attention and MLP compute width $d$. For both scalar and expanded states, let $\xb_l^{\mathrm{in}}\in\RR^d$ denote the vector presented to a standard sublayer, $\mathbf{c}_l=\operatorname{RMSNorm}(\xb_l^{\mathrm{in}})$ its normalized form, and $\mathbf{h}_l=\Fb_l(\mathbf{c}_l)$ the sublayer output. The reported configurations use $\mathbf{h}_l$ as the unnormalized rewrite direction, $\tilde{\kb}_l=\mathbf{h}_l$ and $\kb_l=\tilde{\kb}_l/\|\tilde{\kb}_l\|_2$, and generate the target from the sublayer input and the gate from its normalized form,
\begin{align*}
\vb_l=4\,\sigma\!\left(W_{v,l}\xb_l^{\mathrm{in}}+\mathbf{b}_{v,l}\right),
\qquad
\beta_l=2\sigma\!\left(W_{\beta,l}\mathbf{c}_l+b_{\beta,l}\right),
\end{align*}
where $\sigma$ is the elementwise logistic sigmoid, so each target entry lies in $(0,4)$. Gate logits are computed in fp32, and $b_{\beta,l}$ is initialized to $0$, so every gate starts at $\beta_l\approx1$, the target-match regime.
The ordinary attention or MLP sublayer thus chooses the edit direction, and lightweight branches produce $\vb_l$ and $\beta_l$. These branches add $(d_v+1)(d+1)$ parameters per sublayer, and at the small scale every DDL variant stays within 0.3M parameters of the baseline (Table~\ref{tab:cost_profile}).

\subsection{Scalar residual state: \texorpdfstring{$d_v=1$}{dv=1}}
\label{subsec:experiments_dv1}

For $d_v=1$, $\Xb_l$ reduces to $\xb_l\in\RR^d$, $\xb_l^{\mathrm{in}}=\xb_l$, and Eq.~\eqref{eq:ddl_additive} becomes $\xb_{l+1}=\xb_l+\beta_l(v_l-\kb_l^\top\xb_l)\kb_l$. This setting tests the structured residual update without adding residual value channels or a compressor. The only additions are direction normalization and the lightweight value and gate branches.

\noindent\textbf{Precision-friendly normalization.}
For low-precision training, we implement the unit-direction constraint through RMS normalization and a fixed scale $k_{\text{scale}}=1/\sqrt{d}$: $\hat{\kb}_l=\operatorname{RMSNorm}(\tilde{\kb}_l;\epsilon_k^2/d)$ and $\kb_l=\hat{\kb}_l/\sqrt{d}$. With $\epsilon_k=0$, this is exact $L_2$ normalization. With $\epsilon_k>0$, it is equivalent to $\kb_l=\tilde{\kb}_l/\sqrt{\|\tilde{\kb}_l\|_2^2+\epsilon_k^2}$, so the unit-vector analysis is accurate whenever $\|\tilde{\kb}_l\|_2\gg\epsilon_k$.

\subsection{Expanded residual state: \texorpdfstring{$d_v>1$}{dv>1}}
\label{subsec:experiments_dv_expansion}

For expanded-state DDL, $\Xb_l\in\RR^{B\times T\times d\times d_v}$; we evaluate $d_v=4$. In the no-EC ablations, every value channel starts as the token embedding, $\Xb_0=\xb_{\mathrm{emb}}\mathbf{1}_{d_v}^\top$. Embedding convolution (EC) instead maps each embedding feature to $d_v$ channels with a learnable causal depthwise convolution over tokens~\citep{chollet2017xception},
\begin{equation*}
\Xb_{0,t,i,j}=\sum_{s=0}^{k_e-1} e_{i,j,s}\,\xb_{\mathrm{emb},t-s,i},
\end{equation*}
with kernel size $k_e=4$, initialized to $e_{i,j,0}=1$ and $e_{i,j,s}=0$ for $s>0$ so that EC starts as repetition.

The layer follows a Compress--Process--Rewrite protocol:
\begin{enumerate}[itemsep=1pt, topsep=1pt, leftmargin=*]
\item \textbf{Compress.} Map $\Xb_l$ to $\xb_l^{\mathrm{in}}\in\RR^d$ with the block's compressor $\mathcal{C}_l$, which is applied before the attention sublayer and again, after the attention update, before the MLP sublayer. DDL-TC mixes over the last $k$ tokens and then across value channels; DDL-CC mixes only the value channels of the current token:
\begin{equation*}
\text{TC: }\ \xb_{t,i}^{\mathrm{in}}=\sum_{j=1}^{d_v} r_j\sum_{s=0}^{k-1} c_{i,j,s}\,\Xb_{t-s,i,j},
\qquad
\text{CC: }\ \xb_{t,i}^{\mathrm{in}}=\sum_{j=1}^{d_v} w_{i,j}\,\Xb_{t,i,j}.
\end{equation*}
Both use kernel size $k=4$ and a read vector $\mathbf{r}\in\RR^{d_v}$. In CC, $w_{i,j}=\sum_{a=1}^{d_v} r_a\,c_{i,j-a+\lfloor(k-1)/2\rfloor}$ comes from a zero-padded length-$k$ kernel $c_{i,\cdot}$ along the value axis, with $c_{i,m}=0$ outside ${0\le m<k}$. Kernels start at $1/k$ and $\mathbf{r}$ at $1/d_v$, so TC starts as an average over the last $k$ tokens and all value channels. A separate compressor reads the final state out before the final RMSNorm and the language-model head.
\item \textbf{Process.} Apply the standard width-$d$ attention or MLP sublayer to $\mathbf{c}_l=\operatorname{RMSNorm}(\xb_l^{\mathrm{in}})$, producing $\mathbf{h}_l=\Fb_l(\mathbf{c}_l)$.
\item \textbf{Rewrite.} Set $\tilde{\kb}_l=\mathbf{h}_l$, normalize it to $\kb_l$, generate $\vb_l$ and $\beta_l$ as above, and update $\Xb_l$ using Eq.~\eqref{eq:ddl_additive}.
\end{enumerate}

The read-compare-write operation adds $O(dd_v)$ work and activation traffic per token and layer. The compressor adds $O(kdd_v)$ work for DDL-TC with token-axis kernel size $k$, and $O(dd_v)$ work for DDL-CC. During autoregressive generation, TC keeps the last $k-1$ expanded states at each compressor input and EC the last $k_e-1$ token embeddings, in addition to the attention KV cache; CC needs no token history.

We report four expanded-state variants: DDL-TC and DDL-CC, which enable EC by default, and their ablations DDL-TC w/o EC and DDL-CC w/o EC. DDL-CC is the default expanded-state configuration because it gives the best tradeoff between quality and cost.

\section{Experiments}
\label{sec:experiments}

We compare a nanoGPT-based additive baseline~\citep{Karpathy2022} with scalar DDL ($d_v=1$) and expanded DDL ($d_v=4$). Each configuration uses one training seed and the same 49.15B-token budget.

\begin{figure*}[htp!]
    \centering
    \begin{subfigure}[b]{0.31\linewidth}
        \centering
        \includegraphics[width=\linewidth]{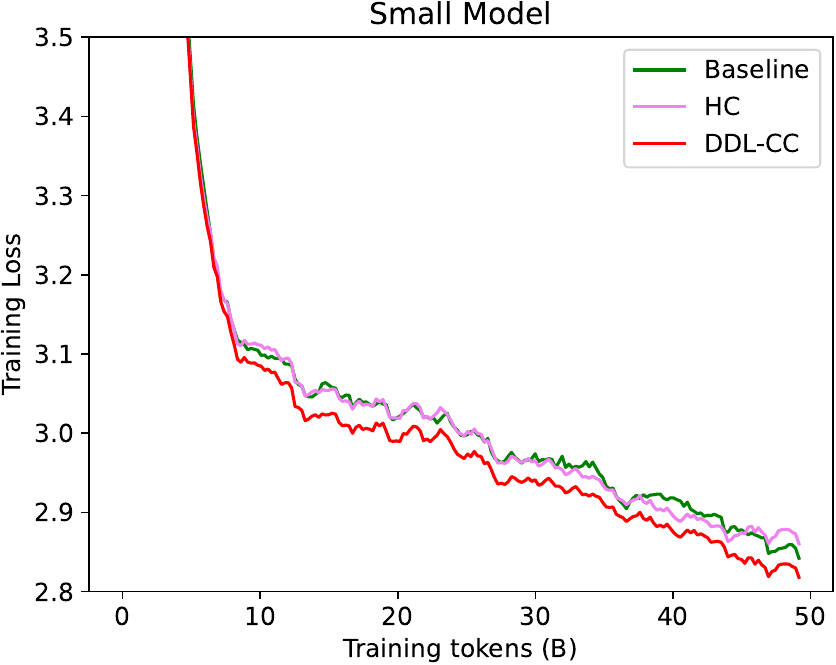}
        \caption{Small Train Loss}
        \label{fig:train_small}
    \end{subfigure}
    \hfill
    \begin{subfigure}[b]{0.31\linewidth}
        \centering
        \includegraphics[width=\linewidth]{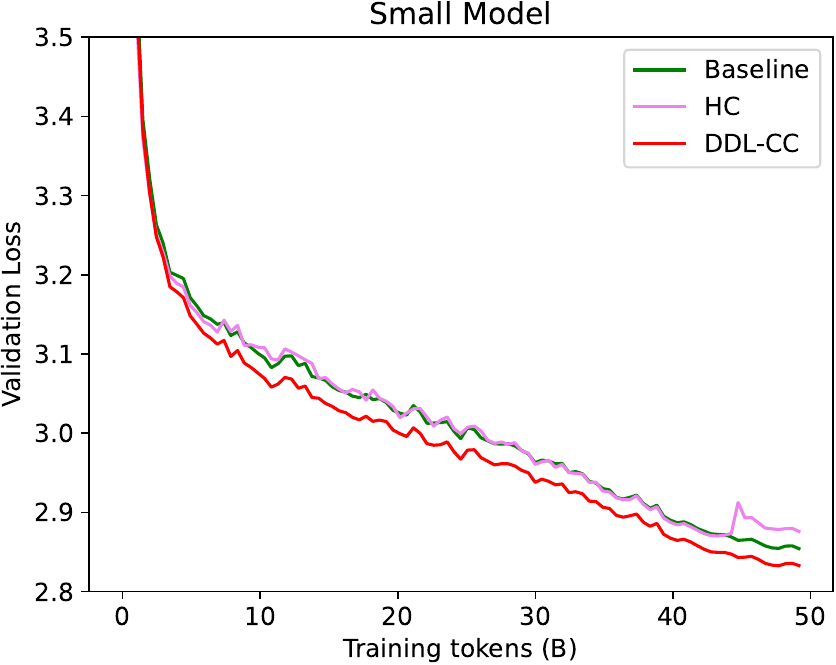}
        \caption{Small Val Loss}
        \label{fig:val_small}
    \end{subfigure}
    \hfill
    \begin{subfigure}[b]{0.31\linewidth}
        \centering
        \includegraphics[width=\linewidth]{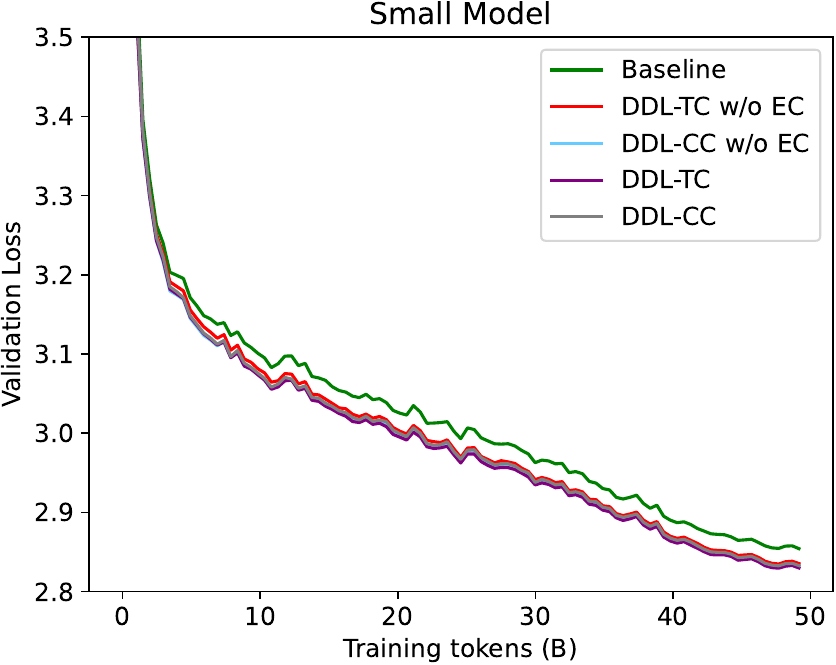}
        \caption{Small DDL Variants Val Loss}
        \label{fig:val_variants}
    \end{subfigure}
    \caption{Small-scale loss curves for the baseline and DDL variants. Panels (a) and (b) also show Hyper-Connections (HC)~\citep{zhuhyper2025} with expansion rate 4, which is not included in the tables.}
    \label{fig:loss_small}
\end{figure*}

\begin{table*}[!ht]
\centering
\caption{One-shot accuracy (\%) for small models. Bold/underlined entries are best/second-best; WG = WinoGrande. Expanded variants use $d_v=4$ and EC unless marked w/o EC.}
\label{tab:small_1_shot}
\resizebox{\textwidth}{!}{%
\begin{tabular}{cccccccccc}
\hline
Model & ARC-C & ARC-E & Hellaswag & OpenBookQA & PIQA & SciQ & Social IQA & WG & Avg. \\ \hline
Baseline & \underline{29.01} & 55.85 & 37.59 & 30.20 & \underline{65.94} & 80.60 & 37.87 & 51.38 & 48.56 \\
DDL $(d_v=1)$ & \textbf{29.35} & 57.49 & 38.08 & 31.80 & 64.85 & 78.50 & 37.77 & \underline{52.01} & 48.73 \\
DDL-TC w/o EC & 27.90 & \underline{58.16} & 38.26 & 30.80 & \textbf{66.49} & 80.30 & \textbf{38.54} & 50.83 & 48.91 \\
DDL-CC w/o EC & 27.82 & \textbf{58.92} & \textbf{38.44} & \underline{33.20} & 65.83 & 79.50 & 38.28 & 51.07 & 49.13 \\
DDL-TC & 28.75 & 57.37 & \underline{38.41} & \textbf{34.40} & 64.47 & \underline{82.00} & 38.38 & \underline{52.01} & \textbf{49.47} \\
DDL-CC & 28.33 & 57.87 & 38.24 & 32.20 & 64.09 & \textbf{82.60} & \underline{38.43} & \textbf{52.57} & \underline{49.29} \\
\hline
\end{tabular}%
}
\end{table*}

\subsection{Experimental settings}
We train on FineWeb-Edu~\citep{lozhkov2024fineweb}. Each run uses 100{,}000 optimization steps, a global batch of 480 sequences, and a sequence length 1{,}024, giving 491{,}520 tokens per update and 49.15B training tokens in total. The models are Llama-style, with RoPE~\citep{su2024roformer}, SwiGLU activations~\citep{shazeer2020glu}, and query/key normalization, and have the depth and width of GPT-2 small and medium (Table~\ref{tab:arch_hyperparam}).

All methods use the same $\mu$P-style parameterization and training recipe~\citep{yang2022tensor}, with no per-method hyperparameter tuning. The learning rate is $1\mathrm{e}{-}3$ with cosine decay and 2{,}000 warmup steps. We use AdamW~\citep{loshchilov2017decoupled} with weight decay $0.1$, $(\beta_1,\beta_2)=(0.9,0.95)$, and gradient clipping at $1.0$. Standard backbone bias terms are disabled and dropout is $0.0$; the DDL gate retains the explicitly initialized output bias described in Section~\ref{sec:ddl_transformer}. Each run uses four NVIDIA H200 GPUs.

\begin{table}[htb!]
\caption{Architecture hyperparameters for the small and medium model sizes.}
\label{tab:arch_hyperparam}
\centering
\small
\begin{tabular}{cccccc}
\toprule
\textbf{Model} & \textbf{\#Param} & \textbf{\#Layer} & \textbf{\#Head} & \textbf{Head Dimension} & \textbf{Hidden Size} \\
\midrule
Small-size Model    & 123.6M             & 12             & 6        & 128                          & 768                          \\
Medium-size Model   & 353.5M             & 24               & 8       & 128                           & 1024                               \\
\bottomrule
\end{tabular}
\end{table}

\subsection{Experimental results}
\begin{figure*}[htp!]
    \centering
    \begin{subfigure}[b]{0.31\linewidth}
        \centering
        \includegraphics[width=\linewidth]{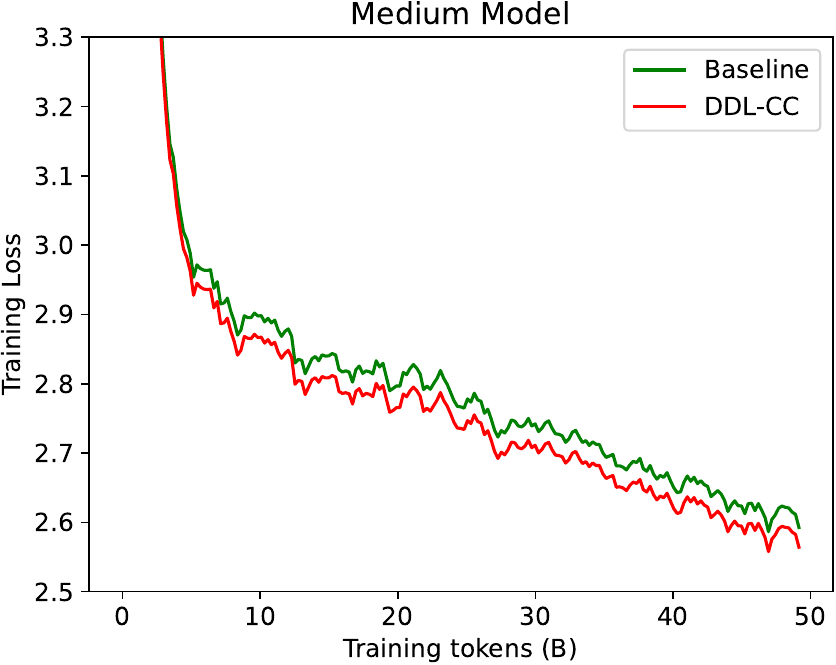}
        \caption{Medium Train Loss}
        \label{fig:train_medium}
    \end{subfigure}
    \hfill
    \begin{subfigure}[b]{0.31\linewidth}
        \centering
        \includegraphics[width=\linewidth]{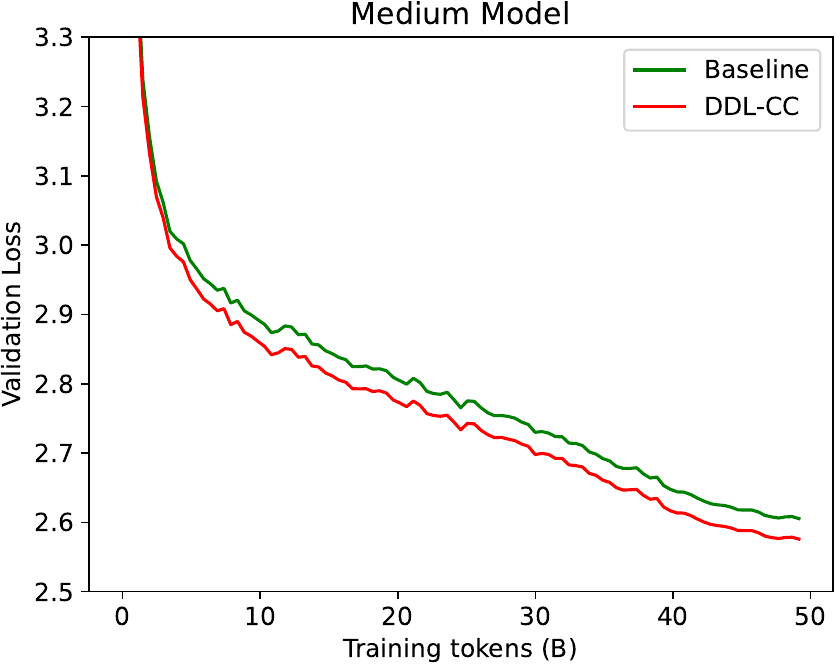}
        \caption{Medium Val Loss}
        \label{fig:val_medium}
    \end{subfigure}
    \hfill
    \begin{subfigure}[b]{0.31\linewidth}
        \centering
        \includegraphics[width=\linewidth]{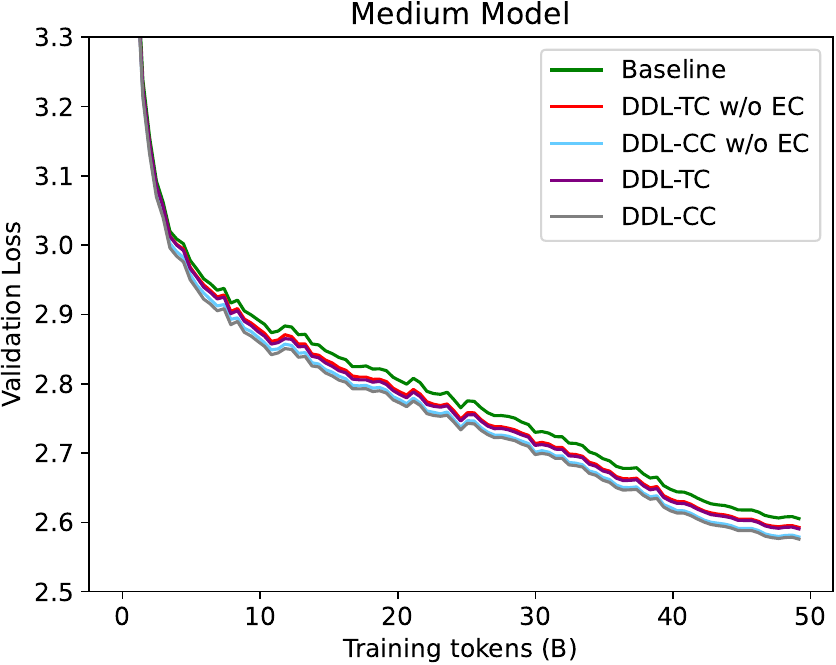}
        \caption{Medium DDL Variants Val Loss}
        \label{fig:val_medium_var}
    \end{subfigure}

    \caption{Medium-scale loss curves for the baseline and DDL variants.}
    \label{fig:loss_medium}
\end{figure*}

\begin{figure*}[htp!]
    \centering
    \begin{subfigure}[b]{0.45\linewidth}
        \centering
        \includegraphics[width=\linewidth]{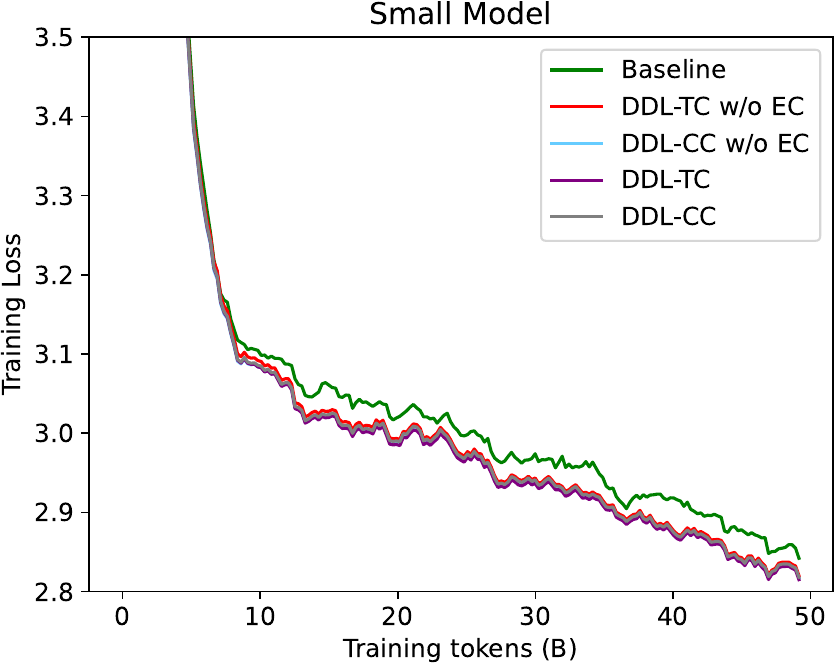}
        \caption{Small DDL Variants Train Loss}
        \label{fig:train_variants}
    \end{subfigure}
    \begin{subfigure}[b]{0.45\linewidth}
        \centering
        \includegraphics[width=\linewidth]{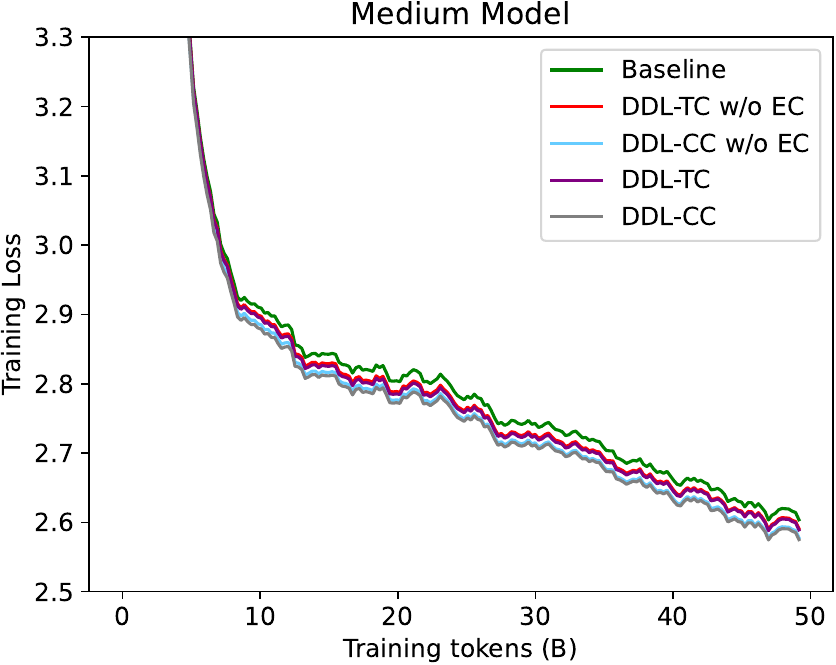}
        \caption{Medium DDL Variants Train Loss}
        \label{fig:train_medium_var}
    \end{subfigure}

    \caption{Training loss curves for the DDL variants at the small and medium scales.}
    \label{fig:variant_train_losses}
\end{figure*}

Figures~\ref{fig:loss_small} and~\ref{fig:loss_medium} show training and validation curves; Figure~\ref{fig:variant_train_losses} gives training curves for the expanded-state variants, and Table~\ref{tab:loss_ppl} reports final validation loss and perplexity. Scalar DDL gives lower final validation loss than the baseline at both scales: $2.8482$ versus $2.8543$ at the small scale and $2.6039$ versus $2.6053$ at the medium scale, corresponding to reductions of $0.0061$ and $0.0014$.

Expanded-state variants give larger reductions. The best small-scale loss is $2.8299$ (DDL-TC), a reduction of $0.0244$ from the baseline, and the best medium-scale loss is $2.5758$ (DDL-CC), a reduction of $0.0295$. These models use $d_v=4$ residual storage and a learned compressor, with embedding convolution enabled unless marked w/o EC. All expanded variants, including those without EC, improve validation loss over the additive baseline.

\paragraph{Matched write-only ablation.}
We isolate the erase term by replacing Eq.~\eqref{eq:ddl_additive} with $\Xb_{l+1}=\Xb_l+\beta_l\kb_l\vb_l^\top$, keeping the state size, EC, compressor, $\kb_l$, $\vb_l$, and $\beta_l$ branches, initialization, data order, and optimizer unchanged. Table~\ref{tab:write_only} reports the small-scale results. Validation loss is 2.8329 for DDL-CC and 2.8406 for its write-only control, a difference of 0.0077; for scalar DDL it is 2.8482 against 2.8608, a difference of 0.0126. Both write-only controls also fall below the additive baseline in one-shot average (47.20 and 47.37 versus 48.56), and the scalar control falls below it in validation loss as well (2.8543 for the baseline). The added branches alone do not reproduce the gain of DDL; the read-and-compare step is needed.

\begin{table}[t]
\centering
\caption{Matched small-scale write-only ablations. Zero-shot and one-shot are eight-task average accuracies (\%). All settings use 100{,}000 steps and the same training-token budget.}
\label{tab:write_only}
\small
\begin{tabular}{lccc}
\toprule
Model & Val. loss & Zero-shot & One-shot \\
\midrule
Baseline & 2.8543 & 47.30 & 48.56 \\
DDL-CC & \textbf{2.8329} & 47.18 & \textbf{49.29} \\
DDL-CC (write-only) & 2.8406 & 46.26 & 47.20 \\
Scalar DDL & 2.8482 & \textbf{47.32} & 48.73 \\
Scalar DDL (write-only) & 2.8608 & 46.97 & 47.37 \\
\bottomrule
\end{tabular}
\end{table}

\paragraph{Hyper-Connections.}
The 4-stream Hyper-Connections (HC) run in Figure~\ref{fig:loss_small} reaches a small-scale validation loss of 2.8364, between DDL-CC (2.8329) and the additive baseline (2.8543). The HC run took 19\,h 56\,min.

We evaluate one-shot and zero-shot performance on ARC~\citep{clark2018think}, HellaSwag~\citep{zellers2019hellaswag}, OpenBookQA~\citep{mihaylov2018can}, PIQA~\citep{bisk2020piqa}, SciQ~\citep{welbl2017crowdsourcing}, Social IQA~\citep{sap2019socialiqa}, and WinoGrande~\citep{sakaguchi2021winogrande} using \texttt{lm-evaluation-harness} \citep{gao2021framework}. Tables~\ref{tab:small_1_shot} and~\ref{tab:medium_1_shot} report one-shot results; Tables~\ref{tab:small} and~\ref{tab:medium} report zero-shot results. Scalar DDL raises the one-shot average by $0.17$ and $0.73$ points at the two scales. The best expanded-state averages improve by $0.91$ and $1.18$ points.

Tables~\ref{tab:cost_profile} and~\ref{tab:cost_profile_medium} report hardware-specific throughput and peak memory. Expanded-state models trade lower validation loss for lower throughput and higher memory use.

\paragraph{Quality--cost tradeoffs.}
At the small scale, scalar DDL retains the baseline's measured peak memory of $2.94$\,GB, but training throughput falls from $1509.6$K to $1330.8$K tokens/s and inference throughput from $1826.1$K to $1605.8$K tokens/s: without residual expansion, the model still pays for normalization and the additional branches. Among expanded models, DDL-TC achieves the lowest validation loss ($2.8299$) and highest one-shot average ($49.47$), whereas DDL-CC gives $2.8329$ and $49.29$ with substantially higher training throughput ($1158.0$K versus $783.5$K tokens/s) and lower peak memory ($3.08$ versus $3.38$\,GB). We therefore use CC as the default.

At the medium scale, DDL-CC has both the lowest validation loss ($2.5758$) and the highest one-shot average ($55.14$) among the reported configurations. Its training throughput is $422.3$K tokens/s, compared with $537.1$K for the baseline and $282.9$K for DDL-TC; peak memory is $7.20$\,GB versus the baseline's $7.06$\,GB. EC also changes the measured execution cost: CC with EC is faster and uses less peak memory than CC w/o EC at both scales in the reported stack. CC and the EC-enabled implementations use optimized Triton kernels.

\paragraph{Effect of input expansion.}
EC's quality effect depends on the compressor and model scale. With TC, enabling EC lowers validation loss from $2.8355$ to $2.8299$ at the small scale and from $2.5927$ to $2.5905$ at the medium scale. With CC, small-scale validation loss instead rises slightly from $2.8321$ to $2.8329$, while the one-shot average increases from $49.13$ to $49.29$; at the medium scale, both metrics improve, from $2.5790$ to $2.5758$ and from $54.92$ to $55.14$.

\begin{table*}[htb!]
\centering
\caption{One-shot accuracy (\%) for medium models. Bold/underlined entries are best/second-best; WG = WinoGrande. Expanded variants use $d_v=4$ and EC unless marked w/o EC.}
\label{tab:medium_1_shot}
\resizebox{\textwidth}{!}{%
\begin{tabular}{cccccccccc}
\hline
Model & ARC-C & ARC-E & Hellaswag & OpenBookQA & PIQA & SciQ & Social IQA & WG & Avg. \\ \hline
Baseline & 33.62 & \textbf{67.05} & 47.42 & 33.20 & \underline{70.24} & 87.30 & 40.28 & 52.57 & 53.96 \\
DDL $(d_v=1)$ & \textbf{35.49} & 65.70 & 46.94 & 34.20 & \textbf{70.35} & 88.90 & \textbf{40.99} & 54.93 & 54.69 \\
DDL-TC w/o EC & 33.02 & \underline{66.16} & 47.83 & 35.60 & 69.86 & \underline{89.50} & \textbf{40.99} & 55.64 & 54.83 \\
DDL-CC w/o EC & 34.30 & 66.08 & \underline{48.65} & \textbf{36.60} & 68.72 & 88.80 & \underline{40.84} & 55.33 & \underline{54.92} \\
DDL-TC & \underline{35.15} & 65.70 & 47.74 & 35.40 & 69.04 & 88.60 & 40.63 & \textbf{56.59} & 54.86 \\
DDL-CC & 34.39 & 65.57 & \textbf{48.92} & \underline{36.00} & 69.48 & \textbf{90.50} & 40.53 & \underline{55.72} & \textbf{55.14}\\
\hline
\end{tabular}%
}
\end{table*}

\begin{table*}[!ht]
\centering
\caption{Zero-shot accuracy (\%) for small models, evaluated with lm-evaluation-harness. Bold/underlined entries are best/second-best; WG = WinoGrande. Expanded variants use $d_v=4$ and EC unless marked w/o EC.}
\label{tab:small}
\resizebox{\textwidth}{!}{%
\begin{tabular}{cccccccccc}
\hline
Model & ARC-C & ARC-E & Hellaswag & OpenBookQA & PIQA & SciQ & Social IQA & WG & Avg. \\ \hline
Baseline & \underline{28.33} & \underline{52.44} & 37.60 & 33.00 & \underline{65.94} & 71.20 & 37.46 & 52.41 & 47.30 \\
DDL $(d_v=1)$ & 26.96 & 52.40 & 37.91 & 32.20 & 64.91 & 72.50 & 38.08 & \textbf{53.59} & 47.32 \\
DDL-TC w/o EC & 27.30 & \textbf{52.95} & 38.40 & \textbf{33.80} & 65.40 & \textbf{73.70} & \underline{38.64} & 50.12 & \underline{47.54} \\
DDL-CC w/o EC & 27.05 & 51.47 & \underline{38.72} & 32.20 & \textbf{66.16} & 71.80 & 38.08 & 51.07 & 47.07\\
DDL-TC & \textbf{28.50} & 51.89 & \textbf{38.82} & \underline{33.20} & 65.29 & 73.00 & \textbf{39.05} & \underline{52.88} & \textbf{47.83} \\
DDL-CC & 27.90 & 51.26 & 38.42 & 31.40 & 64.85 & \underline{73.60} & 37.77 & 52.25 & 47.18 \\
\hline
\end{tabular}%
}
\end{table*}

\begin{table*}[htb!]
\centering
\caption{Zero-shot accuracy (\%) for medium models, evaluated with lm-evaluation-harness. Bold/underlined entries are best/second-best; WG = WinoGrande. Expanded variants use $d_v=4$ and EC unless marked w/o EC.}
\label{tab:medium}
\resizebox{\textwidth}{!}{%
\begin{tabular}{cccccccccc}
\hline
Model & ARC-C & ARC-E & Hellaswag & OpenBookQA & PIQA & SciQ & Social IQA & WG & Avg. \\ \hline
Baseline & 31.74 & \underline{59.85} & 47.91 & 34.00 & 69.21 & 78.10 & \textbf{40.69} & 53.83 & 51.92 \\
DDL $(d_v=1)$ & 32.07 & 59.39 & 47.62 & 34.40 & \textbf{70.08} & 77.30 & 39.61 & 55.01 & 51.94\\
DDL-TC w/o EC & 32.08 & 58.38 & 48.08 & 35.80 & 69.42 & 79.90 & 39.92 & 54.14 & 52.22 \\
DDL-CC w/o EC & 32.08 & \textbf{61.74} & \textbf{49.12} & 34.80 & 69.53 & 80.20 & \underline{40.43} & \underline{55.17} & \textbf{52.88} \\
DDL-TC & \textbf{33.02} & 59.68 & 48.00 & \textbf{36.80} & 68.77 & \underline{81.00} & 39.82 & \textbf{55.88} & \underline{52.87} \\
DDL-CC & \underline{32.59} & 59.55 & \underline{49.01} & \underline{36.00} & \underline{69.70} & \textbf{82.30} & 39.82 & 53.83 & 52.85\\
\hline
\end{tabular}%
}
\end{table*}

\begin{table*}[htb!]
\centering
\caption{Final validation loss and perplexity for small and medium models. The best loss and perplexity in each column are bolded.}
\label{tab:loss_ppl}
\small
\begin{tabular}{ccccc}
\hline
\multicolumn{1}{c}{\multirow{2}{*}{Model}} & \multicolumn{2}{c}{Small}                                     & \multicolumn{2}{c}{Medium}                                     \\ \cline{2-5}
\multicolumn{1}{c}{}                       & Valid Loss & Valid Perplexity & Valid Loss & Valid Perplexity \\ \hline
Baseline                          & 2.8543                            & 17.3616                              & 2.6053                            & 13.5356                              \\
DDL $(d_v=1)$                         & 2.8482                            & 17.2562                              & 2.6039                            & 13.5161                              \\
DDL-TC w/o EC                         & 2.8355                & 17.0381                     & 2.5927                & 13.3654                     \\
DDL-CC w/o EC & 2.8321 & 16.9811 & 2.5790 & 13.1834  \\
DDL-TC & \textbf{2.8299} & \textbf{16.9438} & 2.5905 & 13.3370  \\
DDL-CC & 2.8329 & 16.9947& \textbf{2.5758} & \textbf{13.1420}  \\
\hline
\end{tabular}
\end{table*}

\begin{table*}[!ht]
\centering
\caption{Small-model quality and cost on the same hardware/software stack. Memory factors are relative to the baseline; CC and EC-enabled implementations use optimized Triton kernels. Expanded variants use $d_v=4$.}
\label{tab:cost_profile}
\resizebox{\textwidth}{!}{%
\begin{tabular}{lccccccc}
\toprule
Model & Params & Val loss & Avg eval & Train tok/s (K) & Inference tok/s (K) & Peak memory & Peak-memory factor \\
\midrule
Baseline & 123.6M & 2.8543 & 48.56 & 1509.6 & 1826.1 & 2.94GB & 1.00$\times$ \\
DDL $(d_v=1)$ & 123.6M & 2.8482 & 48.73 & 1330.8 & 1605.8 & 2.94GB & 1.00$\times$ \\
DDL-TC w/o EC & 123.8M & 2.8355 & 48.91 & 673.2 & 688.8 & 3.84GB & 1.31$\times$ \\
DDL-CC w/o EC & 123.7M & 2.8321 & 49.13 & 1019.8 & 1043.0 & 3.47GB & 1.18$\times$ \\
DDL-TC & 123.9M & 2.8299 & 49.47 & 783.5 & 865.1 & 3.38GB & 1.15$\times$ \\
DDL-CC & 123.7M & 2.8329 & 49.29 & 1158.0 & 1220.7 & 3.08GB & 1.05$\times$\\
\bottomrule
\end{tabular}%
}
\end{table*}

\begin{table*}[!ht]
\centering
\caption{Medium-model quality and cost on the same hardware/software stack. Memory factors are relative to the baseline; CC and EC-enabled implementations use optimized Triton kernels. Expanded variants use $d_v=4$.}
\label{tab:cost_profile_medium}
\resizebox{\textwidth}{!}{%
\begin{tabular}{lccccccc}
\toprule
Model & Params & Val loss & Avg eval & Train tok/s (K) & Inference tok/s (K) & Peak memory & Peak-memory factor \\
\midrule
Baseline & 353.5M & 2.6053 & 53.96 & 537.1 & 531.5 & 7.06GB & 1.00$\times$ \\
DDL-TC w/o EC & 354.2M & 2.5927 & 54.83 & 239.7 & 234.6 & 7.68GB & 1.09$\times$ \\
DDL-CC w/o EC & 353.9M & 2.5790 & 54.92 & 358.5 & 337.7 & 7.45GB & 1.06$\times$ \\
DDL-TC & 354.2M & 2.5905 & 54.86 & 282.9 & 291.1  & 7.40GB & 1.05$\times$ \\
DDL-CC & 353.9M & 2.5758 & 55.14 & 422.3 & 400.5 & 7.20GB & 1.02$\times$\\
\bottomrule
\end{tabular}%
}
\end{table*}

\section{Related Work}

\paragraph{Residual and gated pathways.}
Highway Networks~\citep{srivastava2015highway} mix the residual state and a transformed state with coordinate-wise gates; related work explores richer cross-layer and dense residual connections~\citep{chai2020highway,menghani2024laurel,pagliardini2024denseformer,fang2023cross,xiao2025muddformer}. DDL's shortcut $\Ib-\beta_l\kb_l\kb_l^\top$ instead differs from the identity by a rank-1 term: it scales the component along $\kb_l$ by $1-\beta_l$ and leaves the orthogonal complement unchanged, and the same gate scales the target write.

\paragraph{Delta rules and memory updates.}
The delta rule is established in efficient sequence models~\citep{schlag2021linear,yang2024parallelizing}, which update a memory matrix over sequence time; Section~\ref{sec:relation_delta_nets} relates it to DDL term by term. Our contribution is its depth-wise use as the residual interface between Transformer sublayers, and the analysis of that interface. Outer-product memory mechanisms~\citep{mak2025residual} are related through low-rank writes.

\paragraph{Expanded residual states and compression.}
Hyper-Connections~\citep{zhuhyper2025} keep $n$ copies of the residual stream and mix them with learned $n\times n$ matrices, and manifold-constrained Hyper-Connections~\citep{xie2025mhc} restrict that mixing matrix to be doubly stochastic. DDL's $d_v>1$ construction shares their goal of separating persistent state capacity from the width of the expensive sublayers, but its shortcut acts on the feature axis: $\Ib-\beta_l\kb_l\kb_l^\top$ left-multiplies $\Xb_l$ and applies the same operator to every value channel. Its CC compressor is standard learned channel mixing.

\paragraph{Orthogonal and low-rank transformations.}
Householder reflections are classical orthogonal transformations and have been used in neural architectures and adaptation methods~\citep{yang2025path,dong2024efficient,arcas2025hoft}. Other work constrains weights or residual maps to be orthogonal or unitary for stability~\citep{arjovsky2016unitary,jing2017tunable,zhang2021orthogonality,fei2022vit,wang2025otlrm,he2025chaos}. In DDL, the direct shortcut approaches a Householder reflector as $\beta\to2$; this is a limit of the learned gate and no orthogonality constraint is imposed.

\section{Conclusion}

Deep Delta Learning replaces the additive residual update with a delta rule over depth. Each sublayer reads the stream along the direction it writes to and moves that readout toward a target, and one gate spans the range from identity to overwrite. The change is confined to the residual interface, so attention and MLP sublayers are used as they are. In language-model pretraining, DDL improved validation loss and one-shot accuracy over the additive baseline at both scales. Expanding the residual state to four value channels gave the largest gains, and the margin of the best expanded model over the baseline grew from the small to the medium scale. Residual state size is thus a design axis that can be set independently of the compute width of the network, and larger $d_v$, other compressors, and other modalities are natural next steps.

\vspace{5ex}
\bibliographystyle{plainnat}
\bibliography{reference}

\clearpage
\onecolumn
\appendix

\renewcommand{\appendixpagename}{\centering \huge Appendix}
\appendixpage
\counterwithin{theorem}{section}

\startcontents[section]
\printcontents[section]{l}{1}{\setcounter{tocdepth}{2}}
\clearpage

\section{Implementation Details}
\label{app:implementation}
\label{app:expanded_state_impl}

The reported variants are implemented in the following files under \texttt{model/}. Each file also has a Triton implementation with the suffix \texttt{-accelerated}.
\begin{center}
\small
\begin{tabular}{llll}
\toprule
Variant & File & Initial state & Compressor \\
\midrule
DDL ($d_v=1$) & \texttt{DDL-vdim1-gpt-mha-rope-TC.py} & -- & -- \\
DDL-TC w/o EC & \texttt{DDL-gpt-mha-rope-TC.py} & repeated embedding & TC \\
DDL-TC & \texttt{DDL-gpt-mha-rope-TC-EC.py} & EC & TC \\
DDL-CC w/o EC & \texttt{DDL-gpt-mha-rope-CC.py} & repeated embedding & CC \\
DDL-CC & \texttt{DDL-gpt-mha-rope-CC-EC.py} & EC & CC \\
\bottomrule
\end{tabular}
\end{center}
The reported runs use each file's default configuration. The defaults correspond to the settings in Sections~\ref{sec:ddl_transformer}, \ref{subsec:experiments_dv1}, and~\ref{subsec:experiments_dv_expansion} as follows:
\begin{center}
\small
\begin{tabular}{ll}
\toprule
Configuration key & Setting \\
\midrule
\texttt{ddl\_v\_sigmoid}, \texttt{ddl\_v\_sigmoid\_scale} & sigmoid value branch with scale $4$ \\
\texttt{ddl\_beta\_init} & initial gate $\beta_l=1$ \\
\texttt{ddl\_k\_eps} & $\epsilon_k=10^{-5}$ \\
\texttt{input\_embed\_shortconv\_kernel\_size} & EC kernel size $k_e=4$ \\
\texttt{ddl\_state\_shortconv\_kernel\_size} & compressor kernel size $k=4$ \\
\texttt{ddl\_state\_read\_init} & initial read vector $\mathbf{r}=1/d_v$ \\
\bottomrule
\end{tabular}
\end{center}
The code also provides a two-layer $\tanh$ gate, which the reported runs do not use.

\paragraph{Write-only control and checkpoint diagnostics.}
The small-scale write-only controls use 12 Transformer layers, hidden size 768, 100{,}000 updates, 480 sequences of 1{,}024 tokens per global batch, and four H200 GPUs. Both preserve the corresponding DDL model's parameterization, including $\vb_l=4\sigma(W_{v,l}\xb_l^{\mathrm{in}}+\mathbf{b}_{v,l})$, and change only the residual update to $\Xb_{l+1}=\Xb_l+\beta_l\kb_l\vb_l^\top$. For scalar DDL, the state has one value channel; for DDL-CC, it has four.

We inspect a write-only checkpoint on 16 validation documents of 1{,}024 tokens across all 24 sublayers. The mean gate $\beta_l$ exceeds 1.5 in 12 sublayers, and the final sublayer's state column norm is 14.3.

\end{document}